%% file: main.tex
\documentclass[10pt,twocolumn,letterpaper]{article}

\PassOptionsToPackage{table}{xcolor}
\usepackage{cvpr}              

\input{preamble}
\definecolor{cvprblue}{rgb}{0.21,0.49,0.74}
\usepackage[pagebackref,breaklinks,colorlinks,allcolors=cvprblue]{hyperref}

\usepackage[accsupp]{axessibility}
\def\paperID{2496} 
\def\confName{CVPR}
\def\confYear{2026}

\input{math_commands.tex}

\title{UniRefiner: Teaching Pre-trained ViTs to Self-Dispose Dross via Contrastive Register}

\author{Congpei Qiu\textsuperscript{1,4,*} \quad
Zhaoyu Hu\textsuperscript{1,*} \quad
Wei Ke\textsuperscript{1} \quad
Zhuotao Tian\textsuperscript{3,4} \quad
Yanhao Wu\textsuperscript{1} \quad
Tong Zhang\textsuperscript{2,\dag}
\vspace{3pt}
\\
{\textsuperscript{1}Xi'an Jiaotong University, School of Software Engineering} \quad
{\textsuperscript{2}University of Chinese Academy of Sciences} \\
{\textsuperscript{3}Harbin Institute of Technology (Shenzhen)} \quad
{\textsuperscript{4}Shenzhen Loop Area Institute}
}
\begin{document}

\twocolumn[{%
\renewcommand\twocolumn[1][]{#1}%
\maketitle
\vspace{-2em}
\centering
\includegraphics[width=0.9\textwidth]{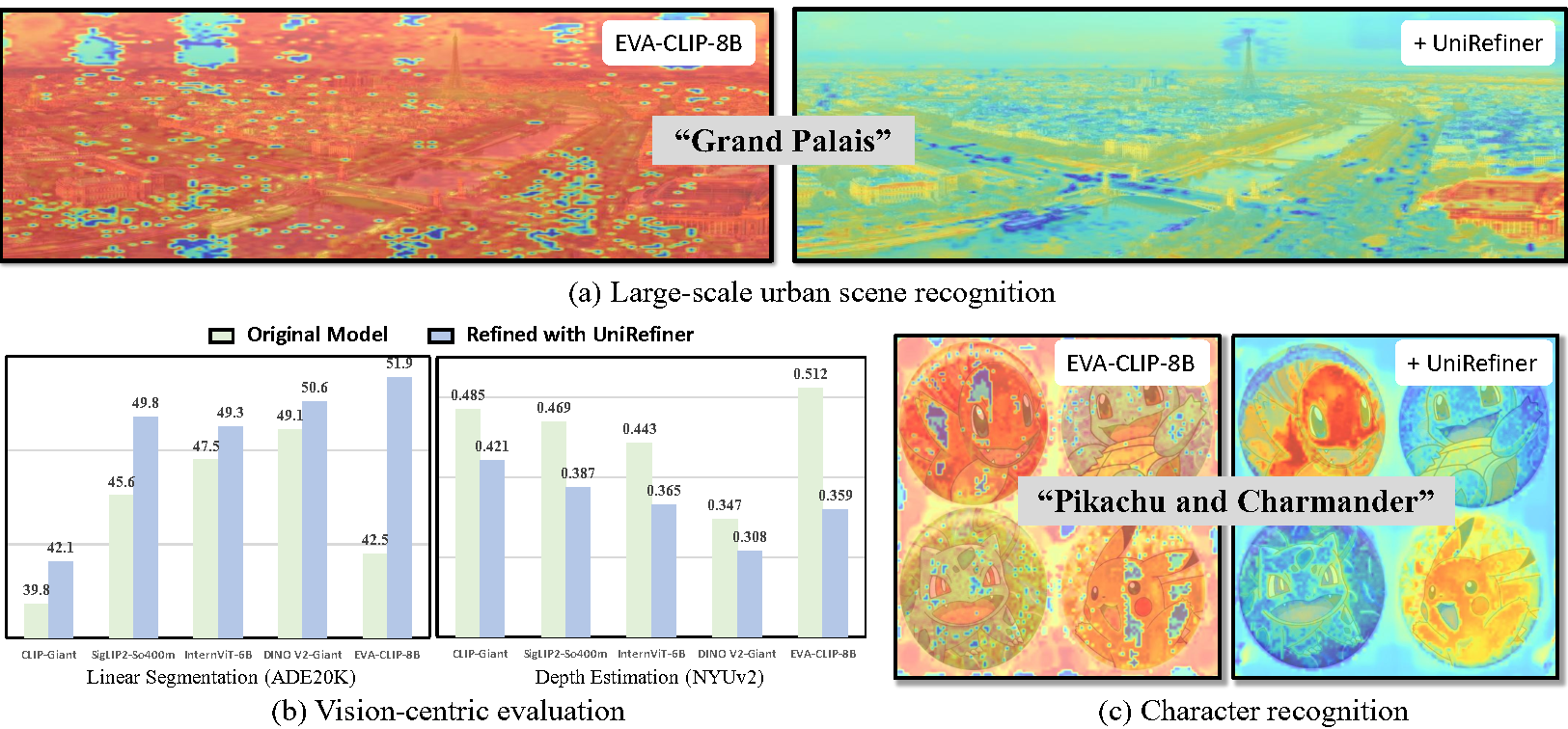}
\vspace{-0.7em}
\captionof{figure}{(a) Visualization of the similarity matrix between language prompts and visual tokens. Warmer colors (red) indicate higher cosine similarity. (a) corresponds to a map of the city of Paris, while (c) depicts characters from the cartoon Pok\'emon. (b) The bar charts present performance comparisons of five large models, with and without our UniRefiner, on segmentation and depth estimation tasks.}
\vspace{1em}
\label{fig:teaser}}]

{\renewcommand{\thefootnote}{} \footnotetext{\textsuperscript{*}Equal contribution. \quad \textsuperscript{\dag}Corresponding author.}}

\input{sec/0_abstract}    
\input{sec/1_intro}

\input{sec/2_related_v2}
\input{sec/3_preliminary_v3}

\input{sec/4_method_v2}

\input{sec/5_experiment_v2}
\input{sec/7_conclusion}
{
    \small
    \bibliographystyle{ieeenat_fullname}
    \bibliography{main}
}


\end{document}

%% file: preamble.tex
\usepackage{multirow}
\usepackage{graphicx}
\usepackage{booktabs}
\usepackage{adjustbox}

%% file: math_commands.tex
\usepackage{amsmath,amsfonts,bm}

\def\eqref#1{equation~\ref{#1}}

\def\1{\bm{1}}

\def\vv{{\bm{v}}}

\def\mA{{\bm{A}}}

\def\mK{{\bm{K}}}

\def\mQ{{\bm{Q}}}

\def\mX{{\bm{X}}}

\def\mZ{{\bm{Z}}}

\DeclareMathAlphabet{\mathsfit}{\encodingdefault}{\sfdefault}{m}{sl}
\SetMathAlphabet{\mathsfit}{bold}{\encodingdefault}{\sfdefault}{bx}{n}

\def\gL{{\mathcal{L}}}

\def\gT{{\mathcal{T}}}

\newcommand{\E}{\mathbb{E}}

\newcommand{\R}{\mathbb{R}}

\newcommand{\softmax}{\mathrm{softmax}}



%% file: sec/0_abstract.tex

\begin{abstract}

Representation learning with Vision Transformers (ViTs) has advanced rapidly, yet the utility of large-scale models in spatially sensitive tasks is hindered by spurious tokens. Prior efforts to mitigate this have been limited, often defining these artifacts narrowly, for example, as simple high-norm outliers. We argue that this scope is insufficient. For dense prediction tasks, we posit that any token failing to encode location-aligned semantics should be treated as a spurious artifact. This broader definition reveals a more complex problem, leading us to systematically categorize and characterize three fundamental types of spurious tokens that corrupt spatial representations. Based on this comprehensive diagnosis, we propose UniRefiner, a universal refinement framework that teaches pre-trained ViTs to self-dispose of these artifacts. UniRefiner uses contrastive registers to explicitly isolate and redistribute spurious tokens via a dual objective: (i) it aligns image tokens with filtered regular tokens to preserve semantics, and (ii) it aligns register tokens with detected spurious tokens to capture the spurious signals. Our method requires only a few epochs of fine-tuning on ~5k images to refine diverse ViTs, including massive models like EVA-CLIP-8B and InternViT-6B. Experiments demonstrate consistent and significant improvements: notably, the refined EVA-CLIP-8B achieves 51.9\% mIoU on ADE20K (+9.4\%), surpassing specialized vision models like DINOv2 (49.1\%), while zero-shot segmentation accuracy improves by up to 22\%. UniRefiner unlocks the latent spatial potential of existing large-scale foundation models, paving the way for their broader application. Project page: \texttt{https://congpeiqiu.github.io/UniRefiner}.
\end{abstract}

%% file: sec/1_intro.tex

\begin{figure}[t]
   \begin{center}
   \includegraphics[width=1\linewidth]{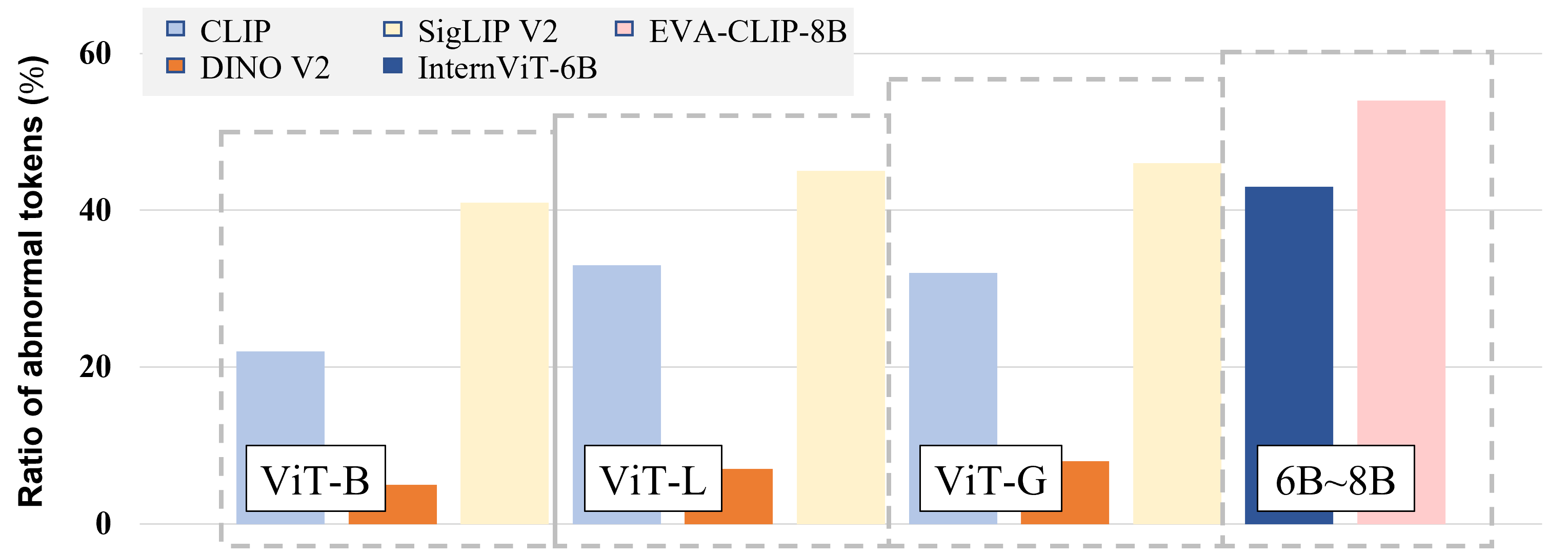}
   \end{center}
    \vspace{-0.6 cm}
   \caption{\textbf{Statistics of spurious token ratios in different ViT models.} We use 1k images sampled from CC3M~\cite{sharma2018conceptual} to calculate the spurious token ratios, including both FP, GP, and AH tokens.
   }
   \label{fig:spurious_statis}
   \vspace{-7 mm}
\end{figure}

\vspace{-5 mm}
\section{Introduction}
\label{sec:intro}








Representation learning based on Vision Transformers (ViTs) has advanced rapidly, yielding notable successes in both vision-centric~\cite{caron2021emerging,oquab2023dinov2,simeoni2025dinov3} and vision–language~\cite{radford2021learning,sun2023eva,chen2024internvl,tschannen2025siglip} domains. This progress has given rise to a flourishing ecosystem of downstream applications in perception~\cite{du2022learning,dong2023maskclip,wu2024dino}, visual understanding~\cite{liu2023visual,zhu2025internvl3}, and image generation~\cite{yu2024representation,xiang2025structured,zheng2025diffusion}, e.g., REPA~\cite{yu2024representation} leverages DINOv2 features to guide the training of generative models, substantially accelerating training while preserving high-quality image synthesis. 

\begin{figure*}[t]
   \begin{center}
   \includegraphics[width=0.95\linewidth]{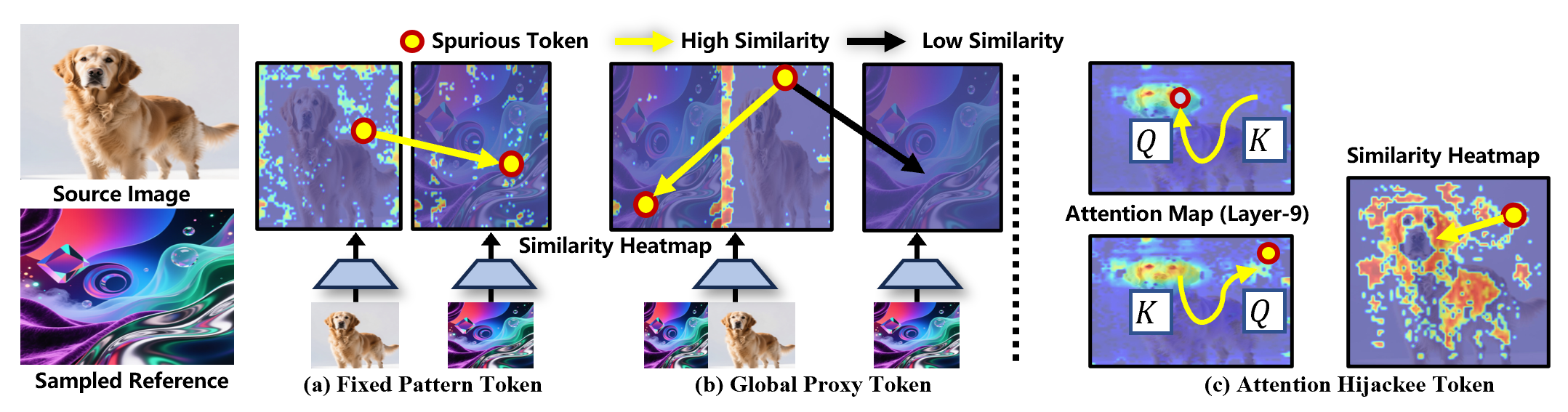}
   \end{center}
    \vspace{-0.8 cm}
   \caption{\textbf{Overview of different spurious token categories.} 
   We sample a source image and a random reference image to illustrate the characteristics of different spurious tokens, highlighting three key categories: \textbf{Fixed Pattern} Tokens: exhibit high cosine similarity with tokens from irrelevant images; 
     \textbf{Global Proxy} Tokens: exhibit high cosine similarity with other tokens within the same feature map but vary across different images;
      \textbf{Attention Hijackee} Tokens: less discriminative tokens dominated by the semantics of more discriminative ones via self-attention. 
   The spurious tokens are marked by \textcolor[HTML]{E3C01E}{yellow dot}. Best viewed in color and zoomed in.
   }
   \label{fig:spurious}
   \vspace{-4 mm}
\end{figure*}

However, a significant bottleneck persists. Despite the wide usage of massive ViT models~\cite{sun2023eva,chen2024internvl}, most downstream methods requiring dense spatial understanding (e.g., segmentation and depth estimation) still default to specialized, vision-only backbones. The intuitive reason is visible in~\cref{fig:teaser}(a)(c): the feature maps of many vision-language large-scale ViTs are spatially inconsistent, contaminated by "spurious tokens" - feature embeddings that are misaligned with their corresponding spatial locations. These spurious tokens severely degrade the representation quality required for dense prediction tasks. Before this spatial potential can be unlocked, we must first understand the problem. Prior studies~\cite{darcet2023vision,wang2024sinder,yang2024denoising}  have noted this issue, often treating spurious tokens as simple high-norm outliers. We find this view is incomplete. In large-scale models, these aberrations are far more diverse and co-exist in large quantities (see~\cref{fig:spurious}). To create an effective solution, we first conducted a systematic analysis of token behaviors that adversely affect representation quality. Our investigation identifies three distinct types of spurious tokens: 1) Fixed Pattern (FP) Tokens, which encode no semantic information and remain consistent across diverse inputs; 2) Global Proxy (GP) Tokens, which capture global context instead of local content; and 3) Attention Hijackee (AH) Tokens, whose local semantics are progressively overwritten by more discriminative neighbors through self-attention. This systematic categorization provides a clear diagnosis. As shown in~\cref{fig:spurious_statis}, these spurious tokens can constitute over 40\% of the features in massive models like EVA-CLIP-8B, making simple outlier removal insufficient.

The above procedure enables a principled algorithm for their identification and mitigation. FP and GP tokens are detected through cosine similarity comparisons against tokens derived from semantically irrelevant references, as both exhibit insufficient spatial coherence with the input image’s content. AH tokens, however, require a distinct approach: since their anomaly arises from pairwise interactions (where the source tokens remain non-anomalous), cosine similarity proves ineffective. Instead, we trace cross-layer attention flow dynamics to identify tokens exhibiting high attention inflow but low outflow—a signature of attention hijacking. This multi-path detection framework forms the \textbf{Spurious Token Filter} that accurately detects and suppresses spurious tokens, ensuring that the retained representations are spatially consistent and semantically meaningful.

\begin{figure}[t]
    \centering       
    \includegraphics[width=1\linewidth]{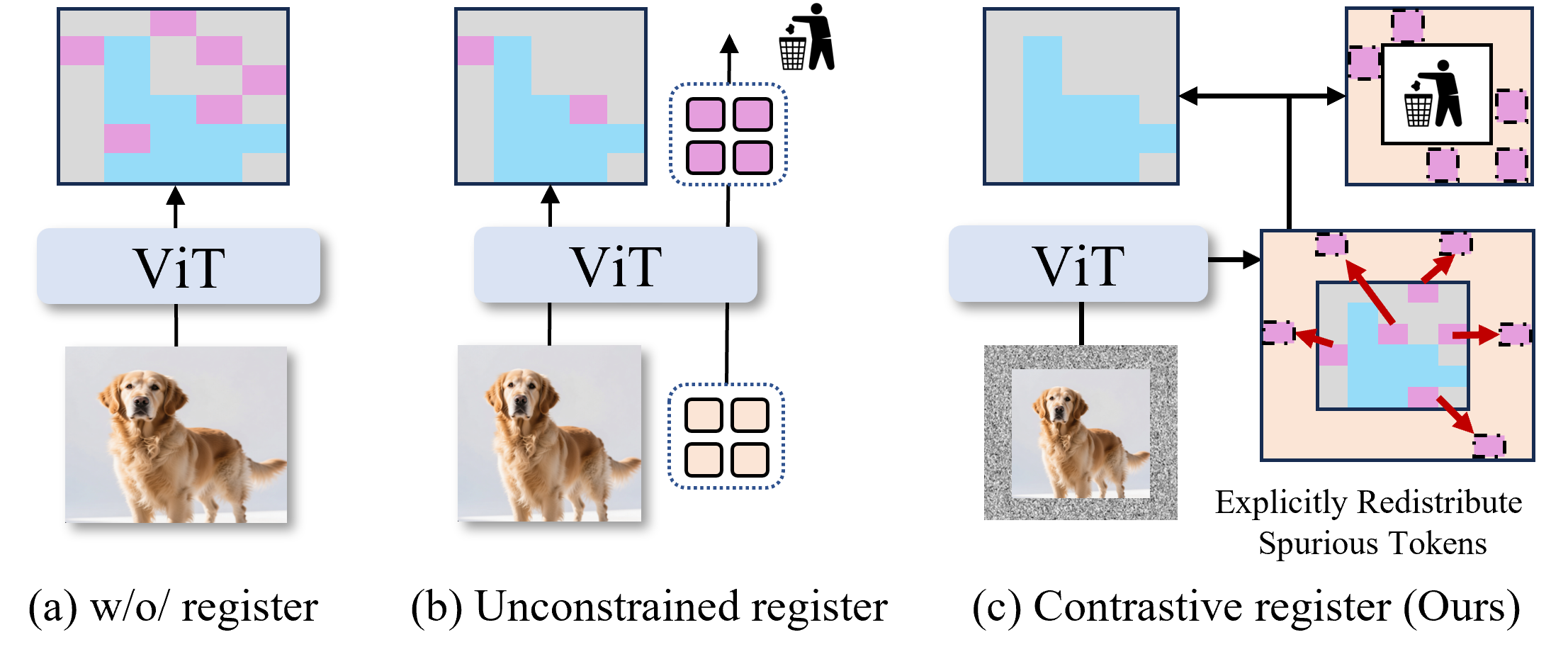}
    \vspace{-0.8cm}
    \caption{\textbf{Register token for spurious token absorption}: (a) ViTs without register tokens are prone to generating spurious tokens; (b) using unconstrained register tokens leads to inefficacy when spurious tokens predominate; (c) our UniRefiner employs explicit learning objectives to concentrate spurious signals.
     }
    \vspace{-0.6cm}
    \label{fig:compare}
\end{figure}
Based on this diagnosis, we propose UniRefiner, a universal, post-hoc framework to refine pre-trained ViTs by teaching them to self-dispose of these artifacts. We leverage register tokens~\cite{darcet2023vision,chen2025vision}, which are appended to the token sequence and discarded post-propagation. However, as shown in~\cref{fig:compare}, the sheer volume of spurious signals in large ViTs means that unconstrained registers are easily overwhelmed. Our key insight is to design a \textbf{Contrastive Register}~(\cref{fig:compare}). Instead of passively hoping registers absorb these artifacts, we explicitly teach the model what to keep and what to discard. We first use our multi-path detection framework to filter the teacher model's output into "regular" and "spurious" sets. Then, UniRefiner enforces a dual-optimization alignment:


(i) Image tokens are aligned with filtered regular tokens to preserve semantic integrity;
(ii) Register tokens are aligned with detected spurious tokens to concentrate spurious signals.
This dual pathway enforces robust diversion of spurious tokens into the register region. To further isolate spurious signals, we introduce a uniformity loss that maximizes divergence between register and image tokens, preventing leakage of corrupted features into the main representation stream. 

The entire UniRefiner framework is implemented via lightweight LoRA~\cite{hu2022lora} fine-tuning and requires no structural modifications to the ViT. It converges in just a few epochs on a small dataset (~5k images) and dramatically enhances the spatial representation quality of large-scale models like EVA-CLIP-8B and InternViT-6B. Our contributions are threefold:

\begin{itemize}
    \item We systematically identify and characterize three fundamental types of spurious tokens in ViTs that impair spatial representation quality.
    \item We propose UniRefiner, a universal post-hoc refinement framework that effectively mitigates spurious tokens through contrastive register learning and self-distillation.
    \item We conduct extensive experiments on vision and vision-language dense prediction tasks, enhancing performance across various dense prediction tasks with notable performance gains (up to 14\% mIoU on semantic segmentation, 22\% mAP on zero-shot semantic segmentation).
\end{itemize}

%% file: sec/2_related_v2.tex
\section{Related Work}
\label{sec:related}

\noindent \textbf{Visual Representation Learning.}~ViT pre-training has advanced rapidly in both vision-centric~\cite{caron2021emerging,xie2022simmim,he2022masked,zhou2021ibot,assran2023self,oquab2023dinov2,simeoni2025dinov3,zhai2023sigmoid} and vision-language~\cite{radford2021learning,jia2021scaling,sun2023eva,tschannen2025siglip,chen2024internvl,maninis2024tips,zhai2023sigmoid} settings. The DINO family~\cite{caron2021emerging,oquab2023dinov2,simeoni2025dinov3} shows strong spatial awareness and is therefore a common backbone for downstream dense applications~\cite{yu2024representation,xiang2025structured,zheng2025diffusion,hong2023lrm}. Recent work has further improved dense and in-context scene understanding with dedicated objectives, including Hummingbird~\cite{balazevic2023towards}, CrIBo~\cite{lebailly2024cribo}, and NeCo~\cite{pariza2025near}. Yet vision-language models such as SigLIPv2~\cite{tschannen2025siglip} and EVA-CLIP~\cite{sun2023eva} still underperform vision-specialized models on spatial tasks. We argue that this gap is largely caused by spurious tokens; once mitigated, these models can rival or even surpass vision-specialized models on dense prediction while retaining multimodal understanding. Unlike prior dense-feature learning approaches, UniRefiner is a lightweight post-hoc refinement framework rather than another dense pretraining stage.


\noindent \textbf{Spurious Tokens in ViTs.}~Large-scale pre-trained Vision Transformers frequently generate spurious tokens that severely degrade spatial representation quality. Literature remains divided on the nature of these tokens~\cite{darcet2023vision,wang2024sinder,yang2024denoising}, ranging from high-norm activations~\cite{darcet2023vision,sun2024massive,wang2024sinder} to position embedding artifacts~\cite{yang2024denoising}---yet no full consensus exists. Consequently, existing mitigation strategies either target singular spurious token types~\cite{darcet2023vision,yang2024denoising,wang2024sinder} or apply identical operations to both spurious and regular tokens~\cite{chen2025vision}, which are inadequate for large ViTs where substantial spuriousness of diverse forms co-exist. To comprehensively address this challenge, we systematically categorize and characterize three fundamental spurious token types, enabling effective identification and targeted mitigation.

%% file: sec/3_preliminary_v3.tex
\section{Preliminary: spurious tokens in ViTs}
\label{sec:preliminary}

For dense prediction, a visual token is useful only if it faithfully represents the content at its own spatial location. We therefore term a token \emph{spurious} whenever its feature misaligns with its corresponding spatial location. In this section, we focus on \emph{how} such tokens behave rather than \emph{why} they emerge. This behavioral perspective is sufficient for the method in Sec.~\ref{sec:methodology}: once the dominant failure patterns are identified, they can be filtered before refinement.

Our empirical study across CLIP~\cite{radford2021learning}, DINOv2~\cite{oquab2023dinov2}, EVA-CLIP~\cite{sun2023eva}, SigLIPv2~\cite{tschannen2025siglip}, and InternViT~\cite{chen2024internvl} reveals three recurring types of spurious tokens. At a high level, they differ as follows: \emph{Fixed Pattern} (FP) tokens remain nearly unchanged across unrelated images; \emph{Global Proxy} (GP) tokens encode scene-level context rather than local evidence; and \emph{Attention Hijackee} (AH) tokens are dominated by more informative neighbors in self-attention. This taxonomy forms the basis of the filter introduced in Sec.~\ref{sec:method:spurious_filter}.

\subsection{Characterizing spurious tokens}
\label{sec:pre:characterize}

Throughout this section, $\mX_s$ denotes a source image and $\mX_{ref}$ an independently sampled reference image. Their feature maps are $\mZ_s$ and $\mZ_{ref}$, and $\Gamma_s,\Gamma_{ref}$ denote the source and reference token sets, respectively.

\noindent \textbf{Fixed Pattern Token.}~FP tokens are the simplest case: they remain nearly unchanged when the image content changes, indicating that they carry little visual-specific information. We identify them by checking whether a source token remains highly similar to \emph{any} token from an unrelated image:
\begin{equation}
    \Gamma_{\text{fp}} = \left\{ i \in \Gamma_s \mid \max_{j \in \Gamma_{ref}} \cos (\mZ_s[i], \mZ_{ref}[j]) \geq \tau_{\text{fp}} \right\}.
\end{equation}
If this condition holds, the token behaves more like a fixed template than a genuine local descriptor.

\noindent \textbf{Global Proxy Token.}~GP tokens are more subtle. Unlike FP tokens, they do change from image to image, yet they still fail to describe their own location because they behave as proxies for the \emph{global} scene context. To expose this behavior, we concatenate $\mX_s$ and $\mX_{ref}$ into a composite image $\mX_{s-ref}$ and extract its feature map $\mZ_{s-ref}$. A source token is identified as a GP token if it is highly similar to tokens from the composite scene, while remaining \emph{inconsistent} with tokens from the standalone reference image:
\begin{equation}
\begin{split}
   \Gamma_{\text{gp}} = \{ i \mid \max_{j} \cos (\mZ_{s-ref}[i], \mZ_{s-ref}[j]) &\geq \tau_{\text{gp}}, \\
    \max_{k} \cos (\mZ_{s-ref}[i], \mZ_{ref}[k]) &< \tau_{\text{fp}}, \\
    \quad i \in \Gamma_s,\; j \in \Gamma_{ref} \}.
    \label{eq:gp}
\end{split}
\end{equation}
The first term captures unusually strong within-scene similarity, while the second rules out the FP case. In other words, GP tokens are not image-invariant; rather, they are scene-level instead of location-level.

\begin{figure*}[t]
   \begin{center}
   \includegraphics[width=0.95\linewidth]{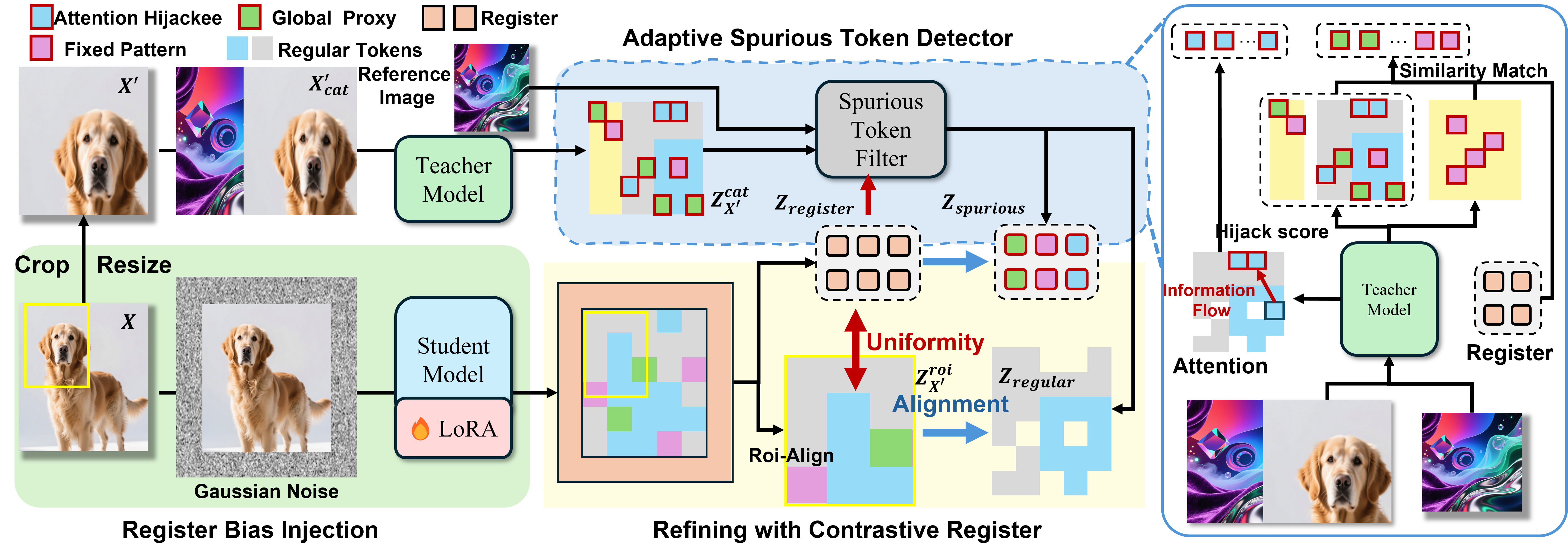}
   \end{center}
    \vspace{-0.7 cm}
   \caption{\textbf{Overview of UniRefiner.} 
    UniRefiner employs the \textit{Spurious Token Filtering} pipeline to identify both regular and spurious tokens (Sec \ref{sec:method:spurious_filter}). Then, we add Gaussian-noise patches as register bias to the input image to obtain image and register tokens, which are respectively aligned with filtered regular and spurious tokens via \textit{Contrastive Register} distillation (Sec \ref{sec:method:contrastive_register}). As distillation progresses, the learned registers will be adopted as adaptive spurious detectors, which further enhance the spurious token filtering process (Sec \ref{sec:method:register_spurious_detector}).
   }
   \label{fig:unirefiner}
   \vspace{-5 mm}
\end{figure*}

\noindent \textbf{Attention Hijackee Token.}~AH tokens cannot be identified by feature similarity alone, because their issue lies in \emph{interaction} rather than static appearance. These tokens are still present in the feature map, but they are rarely used as information sources by other tokens during self-attention. Meanwhile, they can still absorb information from dominant neighbors, so repeated token mixing progressively overwrites their own semantics rather than preserving the content at their spatial locations. We therefore characterize them through asymmetric attention flow and define a \emph{hijack score}:
\begin{equation}
    \begin{aligned}
        \mA^l &= \softmax\!\left(\frac{\mQ^l {\mK^l}^\top}{\sqrt{d}}\right),\\
        h_j   &= \frac{1}{L} \sum_{l} \sum_{i} \mA^l[i,j],
    \end{aligned}
    \label{eq:attn}
\end{equation}
where $h_j$ is the hijack score of token $j$. A small value means that the token is rarely selected as a meaningful source of information, even if it still attends to others; such tokens are therefore likely to be hijackees. We thus define:
\begin{equation}
  \Gamma_{\text{ah}} = \left \{ i \in \Gamma_s \mid h_i < \tau_{\text{ah}} \right \}.
    \label{eq:ah}
\end{equation}

\subsection{Main observation}

The key takeaway is not merely that spurious tokens exist, but that they are both \emph{heterogeneous} and \emph{abundant}. As shown in~\cref{fig:spurious_statis}, larger ViTs tend to exhibit substantially higher spurious-token ratios; for instance, EVA-CLIP-8B contains over 40\% spurious tokens. This explains why previous refinement strategies that target only a single artifact type, or treat all tokens uniformly, can be insufficient for large models. The method in Sec.~\ref{sec:methodology} is built directly on this observation: it first identifies different classes of undesirable tokens, and then explicitly instructs the model which representations to preserve and which ones to divert.

%% file: sec/4_method_v2.tex
\section{Methodology}
\label{sec:methodology}

Our analysis in Sec.~\ref{sec:preliminary} suggests that effective refinement should explicitly distinguish between tokens that preserve location-aligned semantics and those that do not. This distinction is particularly important in large ViTs, where spurious tokens are both diverse and abundant.

To this end, we propose UniRefiner, a universal self-supervised refinement framework with contrastive register, as illustrated in~\cref{fig:unirefiner}. Concretely, UniRefiner first identifies regular and spurious teacher tokens, then trains image tokens to preserve the former and register tokens to absorb the latter, and finally reuses the learned registers as adaptive spurious detectors. We detail these components below.

\subsection{Learning Invariance Against Noise}
UniRefiner follows the representation-invariance principle adopted in previous denoising-based refinement methods~\cite{yang2024denoising,chen2025vision,qiu2025refining}: the noise-free representation of a visual content should remain stable under spatial perturbations. Specifically, for a given image $\mX$ and a pre-trained ViT model $f$, by tracking a visual content $\vv \in \mX$ under different perturbations $\tau \in \gT$, UniRefiner aims to learn:
\begin{equation}
    f_{\theta^*}(\vv, \mX) \rightarrow \E_{\tau \in \gT}[f_{\theta}(\tau(\vv), \tau(\mX))],
\end{equation}
where $\theta$ and $\theta^*$ denote the parameters of the pre-trained and refined ViT models, respectively. Unlike prior approaches that align all tokens across views, UniRefiner first separates regular and spurious tokens, so that refinement is driven by reliable teacher supervision rather than by artifacts.

For refinement, we adopt random resized crop as the spatial perturbation $\tau$, i.e., generating $N$ crops $X'$ for each image $\mX$. To perform self-distillation, a siamese architecture is constructed, where the pre-trained ViT $f_t$ with frozen parameters serves as the teacher branch, and the student branch $f_s$ is initialized from $f_t$, optimized via LoRA~\cite{hu2022lora}.

\subsection{Spurious Token Filter}
\label{sec:method:spurious_filter}
The goal of the spurious token filter is to identify, for each augmented view $\mX'$, which teacher tokens should be preserved and which should be discarded. Following Sec.~\ref{sec:pre:characterize}, we use similarity-based cues for FP/GP tokens and attention-flow cues for AH tokens.

\noindent \textbf{Fixed Pattern Token and Global Proxy Token.}
FP and GP tokens can be exposed by unusually high similarity under irrelevant visual contents. Accordingly, we independently sample a reference image $\mX_{ref}$ and concatenate it with $\mX'$ along a randomly chosen axis (horizontal or vertical), yielding the composite image $\mX'_{cat}$. By feeding both $\mX'_{cat}$ and $\mX_{ref}$ into the teacher branch $f_t$, we obtain their feature maps $\mZ^{\text{cat}}_{\mX'}$ and $\mZ_{ref}$. We then use the same threshold $\tau_{\text{fp-gp}}$ to jointly identify both token types:
\begin{equation}
\begin{split}
\Gamma_{\text{fg-gp}}^{X'} = \{ i \mid &\max_{j} \cos (\mZ^{\text{cat}}_{\mX'}[i], \mZ^{\text{cat}}_{\mX'}[j]) \geq \tau_{\text{fp-gp}} \nonumber \\
&\lor  \max_{k} \cos (\mZ^{\text{cat}}_{\mX'}[i], \mZ_{ref}[k]) \geq \tau_{\text{fp-gp}} \}, \\
\end{split} 
\end{equation}
where $i$ and $j$ index tokens from regions of $\mX'$ and $\mX_{ref}$ in $\mZ^{\text{cat}}_{\mX'}$, respectively, and $k$ indexes tokens from $\mZ_{ref}$. Intuitively, the first term captures GP-like global proxies, while the second captures FP-like cross-image invariance.

\noindent  \textbf{Attention Hijackee Token.}
The identification of AH tokens directly follows Sec.~\ref{sec:pre:characterize}. We store the intermediate attention maps during the forward pass of teacher branch $f_t$ on $\mX'_{cat}$, and compute the hijack score $h_i$ for each token $i$ using Eq.~\ref{eq:attn}. Note that we only consider tokens from the region of $\mX'$ in $\mZ^{\text{cat}}_{\mX'}$ when computing these values.
We then select tokens with relatively low hijack scores using threshold $\tau_{\text{ah}}$:
\begin{equation}
    \Gamma_{\text{ah}}^{X'} = \{ i \mid h_i \leq \mu_h + \tau_{\text{ah}} \sigma_h \},
\end{equation}
where $\mu_h$ and $\sigma_h$ denote the mean and standard deviation of hijack scores within the region of $\mX'$ in $\mZ^{\text{cat}}_{\mX'}$.

Finally, we combine the above index sets to obtain the spurious and regular token sets:
\begin{equation}
    \Gamma_{\text{spu}}^{X'} = \Gamma_{\text{fg-gp}}^{X'} \cup \Gamma_{\text{ah}}^{X'}, \quad \Gamma_{\text{regu}}^{X'} = \bar{\Gamma}_{\text{spu}}^{X'}.
\end{equation}

\subsection{Contrastive Register}
\label{sec:method:contrastive_register}

Unlike previous constraint-free register-based methods~\cite{darcet2023vision,chen2025vision}, we argue that explicit constraints are necessary, especially for large ViTs with abundant spurious tokens. In particular, refinement should specify not only which representations to preserve, but also where the undesired signals should be redirected.

\noindent  \textbf{Register Bias Injection.}~We append Gaussian-noise patches around the input image to serve as post-hoc register bias with predefined register factor $r_{reg}$. Specifically, given a feature map $\mZ \in \R^{H \times W \times D}$, the injected feature map is in $\R^{(H+2N_{reg}) \times (W+2N_{reg}) \times D}$, where $N_{reg} = \lceil \frac{\min(W, H)}{r_{reg}} \rceil$.

This design offers three advantages: 1) it integrates seamlessly with pre-trained models without any architectural modification; 2) it remains flexible across diverse input resolutions; 3) Gaussian noise introduces randomness, preventing the collapse of register tokens.

\renewcommand{\arraystretch}{1.2}
\begin{table*}[t!]
\centering
\caption{\textbf{Linear probing  evaluation results on segmentation and depth.} For semantic segmentation, we report the mean Intersection over Union (mIoU, \%) metric and mean accuracy (mAcc, \%). For monocular depth estimation, we report Root Mean Squared Error (RMSE), absolute relative error (Abs Rel), and accuracy under threshold $\delta_1$.}
\vspace{-2 mm}

\setlength{\tabcolsep}{4.5pt}
 
\resizebox{0.95\textwidth}{!}{
\begin{tabular}{l|c|cc|cc|cc|ccc}
\hline
\multirow{2}{*}{Method} & \multirow{2}{*}{ViT} & \multicolumn{2}{c|}{ADE20k} & \multicolumn{2}{c|}{CityScapes} & \multicolumn{2}{c|}{ Pascal VOC} & \multicolumn{3}{c}{NYUd} \\ 
                        && mIoU($\uparrow$) & mAcc($\uparrow$) & mIoU($\uparrow$) & mAcc($\uparrow$) & mIoU($\uparrow$) & mAcc($\uparrow$) & RMSE($\downarrow$) & Abs Rel($\downarrow$) & $\delta_{1}$($\uparrow$) \\ \hline
DINOv2~\cite{oquab2023dinov2} & G/14 & 49.1&61.3&71.5&81.1&84.2&90.7&0.347&0.097&91.9 \\
\rowcolor[HTML]{EFEFEF} 
DINOv2 + UniRefiner & G/14 &\textbf{50.6} & \textbf{63.0} & \textbf{73.4} & \textbf{83.0} & \textbf{85.4} & \textbf{92.2} & \textbf{0.308} & \textbf{0.085} & \textbf{94.1} \\ \hline
CLIP~\cite{radford2021learning} & G/14 &39.8&52.4&58.6&68.2&74.6&83.8&0.485& 0.147 & 81.3 \\
\rowcolor[HTML]{EFEFEF} 
CLIP + UniRefiner & G/14 & \textbf{42.1} & \textbf{54.9} & \textbf{61.7} & \textbf{71.5} & \textbf{76.7} & \textbf{86.0} & \textbf{0.421} & \textbf{0.125} & \textbf{85.9} \\ \hline
InternViT~\cite{chen2024internvl} & 6B/14 &47.5&59.5&70.8&79.9&81.0&88.7&0.443&0.130&84.4\\
\rowcolor[HTML]{EFEFEF} 
InternViT + UniRefiner & 6B/14 &\textbf{49.3} & \textbf{61.3} & \textbf{71.8} & \textbf{80.9} & \textbf{83.1} & \textbf{90.2} & \textbf{0.365} & \textbf{0.102} & \textbf{90.4} \\ \hline
SigLIPv2~\cite{tschannen2025siglip} & So/16 &45.6&58.6&63.4&74.6&77.8&85.3&0.469&0.143&82.0\\
\rowcolor[HTML]{EFEFEF} 
SigLIPv2 + UniRefiner & So/16 &\textbf{49.8} & \textbf{62.9} & \textbf{67.9} & \textbf{78.5} & \textbf{82.1} & \textbf{89.9} & \textbf{0.387} & \textbf{0.112} & \textbf{88.6} \\ \hline
EVA-CLIP~\cite{sun2023eva} &8B/14& 42.5&55.2&69.5& 78.4&69.6&79.4&0.512&0.161&77.9\\
\rowcolor[HTML]{EFEFEF} 
EVA-CLIP + UniRefiner &8B/14 &\textbf{51.9} & \textbf{64.9} & \textbf{74.6} & \textbf{83.1} & \textbf{83.6} & \textbf{91.4} & \textbf{0.359} & \textbf{0.102} & \textbf{90.9} \\ \hline
\end{tabular}%
}
\vspace{-.2cm}
\label{tab:linear_probe}
\end{table*}

\begin{figure*}[t]
    \centering
    \includegraphics[width=0.9\linewidth]{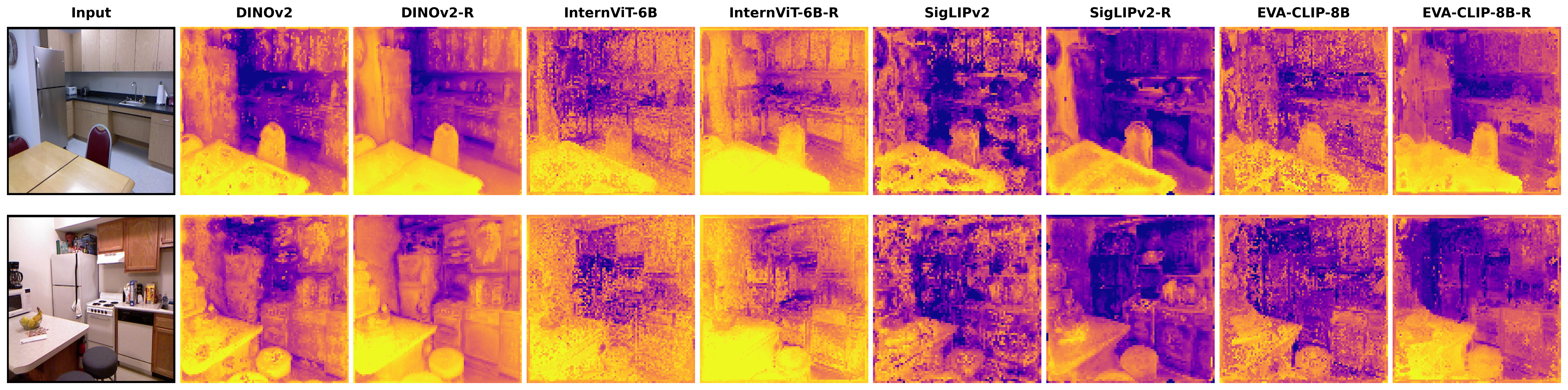}
    \caption{\textbf{Depth estimation comparison on NYUv2.} Predicted depth maps from linear probing on frozen backbone features, shown before and after UniRefiner refinement (suffix ``-R''). UniRefiner produces smoother depth maps with fewer spurious artifacts and improved boundary preservation.}
    \label{fig:depth_vis}
    \vspace{-4 mm}
\end{figure*}

\noindent  \textbf{Contrastive Distillation.}~
The contrastive objective separates the preservation of regular semantics from the absorption of spurious signals. By forwarding register-injected image $\mX$ through the student branch, we obtain ViT tokens $\mZ_{inj}$, which contain image region $\mZ$ and register region $\mZ_{reg}$.
We use ROI-Align~\cite{he2017mask} to extract the feature map aligned with the crop $\mX'$ from $\mZ$, denoted as $\mZ_{\mX'}^{\text{roi}}$. Consequently, we respectively align $\mZ_{\mX'}^{\text{roi}}$ and $\mZ_{reg}$ with identified regular tokens and spurious tokens from the teacher branch:
{\small
\begin{equation}
    \label{eq:loss_nm}
    \gL_{\text{regu}} = \frac{1}{| \Gamma_{\text{regu}}^{X'} |} \sum_i \gL_{NCE}(\mZ_{\mX'}^{\text{roi}}[i], \mZ^{\text{cat}}_{\mX'}[i]), \ i \in \Gamma_{\text{regu}}^{X'},
\end{equation}
}
\begin{equation}
    \label{eq:loss_ab}
    \begin{split}
    \gL_{\text{spu}} = \frac{1}{| \mZ_{reg} |} \sum_i \gL_{NCE}(\mZ_{reg}[i], \mZ^{\text{cat}}_{\mX'}[l]), \\
    l = \arg\max_{j \in \Gamma_{\text{spu}}^{X'}} \cos(\mZ_{reg}[i], \mZ^{\text{cat}}_{\mX'}[j]),
    \end{split}
\end{equation}
where $\gL_{NCE}$ denotes the InfoNCE loss~\cite{oord2018representation}. Eq.~\ref{eq:loss_nm} encourages student image tokens to preserve location-aligned teacher semantics. Eq.~\ref{eq:loss_ab} assigns each register token to its most similar spurious teacher token, thereby encouraging the student ViT to redirect spurious information into the register region. We further apply the Uniformity term~\cite{wang2020understanding} in InfoNCE to separate image and register representations:
\begin{equation}
    \label{eq:loss_u}
    \gL_{\text{uni}} = \frac{1}{| \mZ_{\mX'}^{\text{roi}} |} \sum_i \log \sum_{j} \exp(\frac{{\mZ_{\mX'}^{\text{roi}}[i]}^T \mZ_{reg}[j]}{\tau_{\text{uni}}}),
\end{equation}
which further purifies image tokens.
The overall objective is:
\begin{equation}
    \label{eq:loss_unirefiner}
    \gL_{\text{UniRefiner}} = \gL_{\text{regu}} + \lambda_{\text{spu}} \gL_{\text{spu}} + \lambda_{\text{uni}} \gL_{\text{uni}},
\end{equation}
where $\lambda_{\text{spu}}$ and $\lambda_{\text{uni}}$ are balancing factors.

\subsection{Registers as Spurious Detector}
\label{sec:method:register_spurious_detector}

As refinement progresses, the fine-tuned student ViT gradually migrates spurious tokens into the register region, yielding registers that are increasingly enriched with spurious content. We therefore leverage these learned registers as adaptive spurious detectors to further enhance the filtering process in Sec.~\ref{sec:method:spurious_filter}. Specifically, for register tokens $\mZ_{reg}$, tokens in $\mZ^{\text{cat}}_{\mX'}$ are marked as spurious if their cosine similarity with any register token exceeds a threshold $\tau_{\text{reg}}$:
\begin{equation}
\Gamma_{\text{reg}}^{X'} = \{ i \mid \max_{j} \cos (\mZ^{\text{cat}}_{\mX'}[i], \mZ_{reg}[j]) \geq \tau_{\text{reg}} \},
\end{equation} 
which is adopted to complement previous spurious sets.

%% file: sec/5_experiment_v2.tex
\section{Experiment}
\label{sec:experiment}

\begin{table*}[t]
    \renewcommand{\arraystretch}{0.88}
    \centering
    \caption{Results on open-vocabulary semantic segmentation based on VLM features.}
    \begin{adjustbox}{width=0.9\textwidth}
    \begin{tabular}{l|c|ccc|ccccccc|c}
        \toprule
        \multirow{2}{*}{Method} &\multirow{2}{*}{ViT}& \multicolumn{3}{c}{With a background category} & \multicolumn{5}{c}{Without background category} & \multirow{2.5}{*}{Avg} \\
        \cmidrule(lr){3-5} \cmidrule(lr){6-10}
        & & VOC21  & Context60 & COCO-Obj & VOC20 & CityScape & Context59 & ADE & COCO-Stf &  \\
        \midrule
        CLIP~\cite{radford2021learning} &G/14&42.7&22.9&20.9&56.0&26.9&25.1&16.2&18.2&28.6 \\
        \rowcolor[HTML]{EFEFEF}
        CLIP + UniRefiner & G/14 &\textbf{45.7} & \textbf{25.0} & \textbf{21.7} & \textbf{57.3} & \textbf{31.8} & \textbf{28.1} & \textbf{21.6} & \textbf{19.6} & \textbf{31.4} \\ \midrule
        SigLIPv2~\cite{tschannen2025siglip} & So/16 & 35.5 &21.4 & 21.1  & 68.0 & 23.6 &24.2  & 16.3  & 16.8  & 28.4 \\
        \rowcolor[HTML]{EFEFEF}
        SigLIPv2 + UniRefiner & So/16 &\textbf{51.3} & \textbf{29.4} & \textbf{26.9} & \textbf{69.1} & \textbf{34.6} & \textbf{34.3} & \textbf{21.7} & \textbf{23.3} & \textbf{36.3} \\  \midrule
        EVA-CLIP~\cite{sun2023eva} & 8B/14 & 24.5 & 14.1& 13.5  &70.6  & 11.4 &16.0  & 11.7  & 10.8  & 21.6 \\
        \rowcolor[HTML]{EFEFEF}
        EVA-CLIP + UniRefiner & 8B/14 &\textbf{46.4} & \textbf{26.4} & \textbf{23.8} & \textbf{78.2} & \textbf{30.1} & \textbf{29.8} & \textbf{19.2} & \textbf{20.5} & \textbf{34.3} \\
        \bottomrule
    \end{tabular}
    \end{adjustbox}
\label{tab:zero-shot}
\end{table*}

\begin{figure*}[t]
   \begin{center}
   \includegraphics[width=0.9\linewidth]{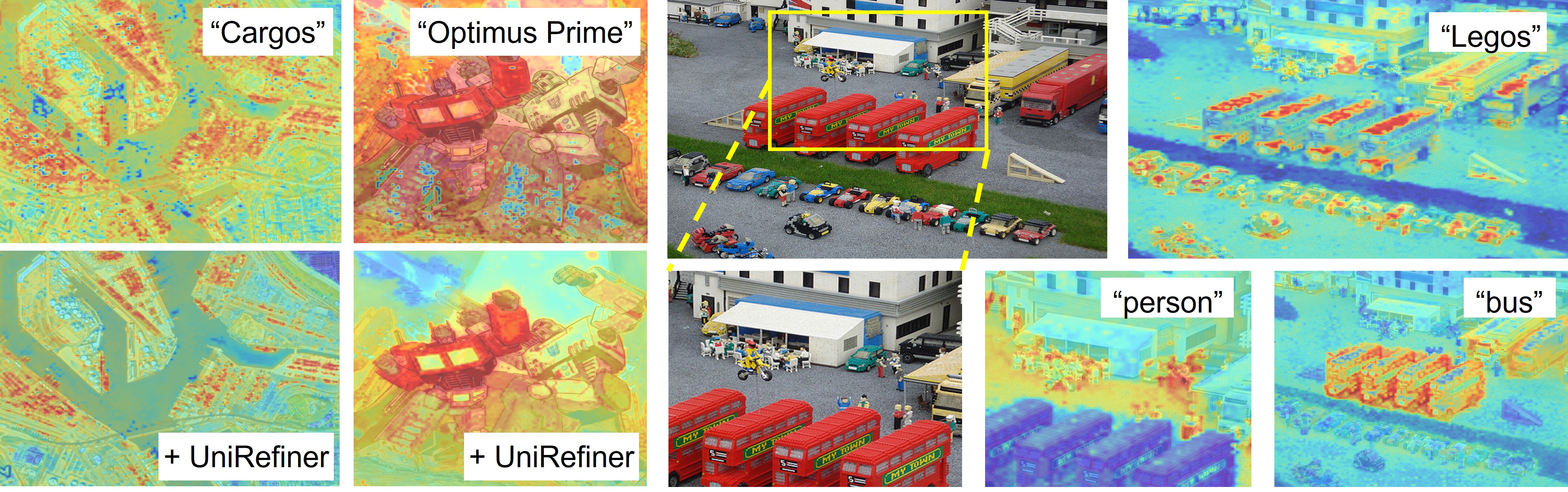}
   \end{center}
    \vspace{-0.7 cm}
   \caption{\textbf{Heatmaps visualization of cosine similarity between image and text embeddings under high resolution.} \textbf{(left)} we compare vanilla and UniRefiner-refined EVA-CLIP-8B, with text requiring both localization and world knowledge; \textbf{(right)} we visualize refined model on a complex visual scene. We upsample the input image of both visualizations to a high resolution of \textbf{1792×1792}, leading to feature maps of \textbf{128×128} tokens. Best viewed in color and zoomed in.
   }
   \label{fig:visual_vl}
   \vspace{-4 mm}
\end{figure*}

In this section, we comprehensively evaluate the generalization ability and effectiveness of UniRefiner on various models and downstream tasks requiring dense spatial representations. We first detail the implementation specifics in Sec.~\ref{sec:experiment:implementation}, then evaluate ViT backbones refined by UniRefiner on vision-centric dense prediction tasks in Sec.~\ref{sec:experiment:vision_dense}, vision-language dense prediction tasks in Sec.~\ref{sec:experiment:vl_dense}, and provide an application example of image generation in Sec.~\ref{sec:experiment:generation}. Finally, we conduct ablation studies and further analysis in Sec.~\ref{sec:experiment:ablation}. Full details are in the supplementary material.

\subsection{Implementation}
\label{sec:experiment:implementation}

We apply UniRefiner to a variety of pre-trained ViT backbones, including vision-centric models (DINOv2~\cite{oquab2023dinov2}) and vision-language models (EVA-CLIP~\cite{sun2023eva}, CLIP~\cite{radford2021learning}, InternViT~\cite{chen2024internvl}, SigLIPv2~\cite{tschannen2025siglip}), spanning 400M–8B parameters. For each backbone we fine-tune only LoRA adapters~\cite{hu2022lora} (rank 8). Our framework converges rapidly: two epochs on a random 5k subset of CC3M~\cite{sharma2018conceptual} suffice even for large models such as EVA-CLIP-8B and InternViT-6B. As examples, training takes $\sim$5 minutes for SigLIPv2-So400m on 4×H100 and $\sim$20 minutes for EVA-CLIP-8B on 8×H100. Full implementation and training details are provided in the supplementary material.

\subsection{Vision-centric Dense Prediction}
\label{sec:experiment:vision_dense}

\noindent \textbf{Datasets and Tasks.}~We evaluate UniRefiner on two fundamental vision-centric dense prediction tasks: semantic segmentation and monocular depth estimation. For semantic segmentation, we follow the linear probing protocol~\cite{oquab2023dinov2}, where a single linear layer is trained to predict pixel-wise semantic classes from frozen backbone features.
Experiments are conducted on ADE20K~\cite{zhou2017scene}, CityScapes~\cite{cordts2016cityscapes}, and PASCAL VOC 2012~\cite{everingham2015pascal}.
For depth estimation, we follow~\cite{oquab2023dinov2} to evaluate on the NYUv2-Depth dataset~\cite{silberman2012indoor} using a linear evaluation protocol.

\noindent \textbf{Experimental Results.}~Table~\ref{tab:linear_probe} reports linear dense-prediction results for various ViT backbones. UniRefiner consistently and substantially improves performance across pre-training paradigms and model scales. Notably, the refined SigLIPv2-So achieves 49.8\% mIoU on ADE20K, outperforming the larger DINOv2 Giant despite its smaller size. Remarkably, refined EVA-CLIP-8B—originally one of the weaker baselines—reaches 51.9\% and 74.6\% mIoU on ADE20K and CityScapes, respectively, indicating its potential was largely suppressed by spurious tokens. The same trend holds for depth estimation and is even more pronounced (visualized in~\cref{fig:depth_vis}), as depth prediction places greater emphasis on spatial integrity. Overall, these results demonstrate UniRefiner's effectiveness and suggest that the spatial capabilities of some large-scale vision–language models have been substantially underestimated.

\subsection{Vision-language Dense Prediction}
\label{sec:experiment:vl_dense}

\noindent \textbf{Datasets and Tasks.}~To evaluate whether UniRefiner maintains the vision-language alignment capabilities of pre-trained models while enhancing spatial representations, we assess on zero-shot open-vocabulary semantic segmentation.
The evaluation protocol follows MaskCLIP~\cite{zhou2022extract}, which directly leverages the similarity matrix between image patch embeddings and text embeddings to produce pixel-wise predictions, with only training-free modification on the final attention layer. We evaluate on eight benchmarks: PASCAL VOC (VOC20/21), PASCAL Context (Context59/60)~\cite{mottaghi2014role}, Cityscapes, ADE20K, COCO-Object and COCO-Stuff164k~\cite{caesar2018coco}.


\noindent \textbf{Experimental Results.}~As shown in Table~\ref{tab:zero-shot}, the performance gains are consistent with the improvement in spatial integrity, reaching up to +19\% mIoU on Cityscapes for EVA-CLIP-8B and nearly tripling its original performance. This indicates that UniRefiner preserves the vision-language alignment capabilities of pre-trained models while further benefiting from enhanced spatial representations. As a further investigation, we ask whether the world knowledge learned by large-scale vision-language models can be transferred to dense prediction after refinement. We therefore provide high-resolution visualizations of similarity heatmaps between image and text embeddings under challenging scenarios, including remote sensing, character recognition, and complex visual scenes, as shown in~\cref{fig:visual_vl}. These results demonstrate that the refined EVA-CLIP-8B has strong potential for a broader range of applications requiring both spatial understanding and world knowledge.


\begin{table}[t]
    \centering
    \caption{\textbf{Comparison of image generation} using REPA on ImageNet 256×256 with CFG. We apply guidance interval for all models following~\cite{yu2024representation}. '-R' refers to refined with UniRefiner.}
    \vspace{-2 mm}
        \label{tab:generation}
    \begin{adjustbox}{width=0.4\textwidth}
 \begin{tabular}{ccccccc}
\hline
Model & Iteration & FID $\downarrow$ & sFID $\downarrow$ & IS $\uparrow$ \\ \hline
SiT-XL/2+DINOv2 & 400K & 2.02 & \textbf{4.42} & 272.4 \\ \hline
SiT-XL/2+SigLIPv2 & 400K & 2.21 & 4.43 & 266.3 \\
\rowcolor[HTML]{EFEFEF}  SiT-XL/2+SigLIPv2-R & 400K & \textbf{1.96} & 4.43 & \textbf{276.2} \\ \hline
\end{tabular}
\end{adjustbox}
\vspace{-3 mm}
\label{tab:image-generation}
\end{table}

\subsection{Application: Image Generation}
\label{sec:experiment:generation}
We provide an example application of UniRefiner-refined ViT, following REPA~\cite{yu2024representation}, the recent advance directly supervises the hidden layers of diffusion models with frozen ViT features to accelerate training. We replace the DINOv2 backbone in REPA with vanilla and refined SigLIPv2 models, and evaluate the generation quality as in Table~\ref{tab:generation}, respectively. We follow the evaluation protocol of REPA, using CFG guidance with a guidance interval of $[0,0.7]$.
Given the strong spatial awareness of refined SigLIPv2 features, our UniRefiner-enhanced model outperforms both the vanilla SigLIPv2 and DINOv2 counterparts, producing images with higher fidelity and better structure consistency.

\subsection{Ablation Study and Analysis}
\label{sec:experiment:ablation}
In this section, unless otherwise specified, we use SigLIP2-So400m as the backbone and linear probing on ADE20K as the quantitative indicator.


\begin{figure}[t]
   \begin{center}
   \includegraphics[width=0.9\linewidth]{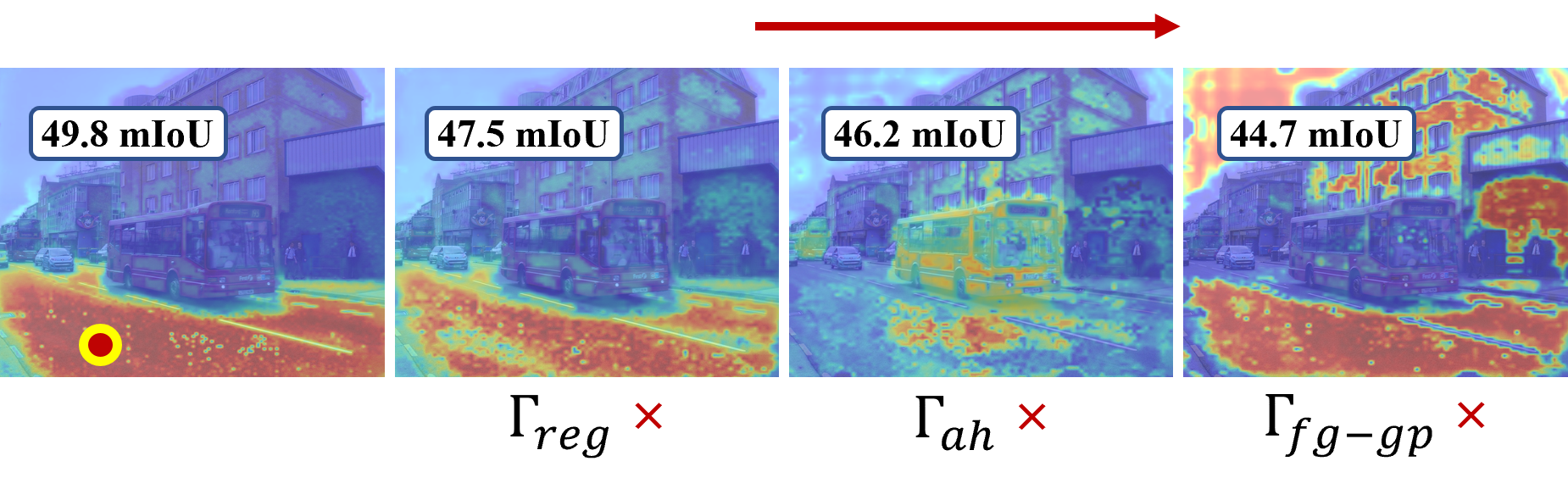}
   \end{center}
    \vspace{-0.8 cm}
\caption{\textbf{Ablation of the spurious token filter.} We gradually ablate each filtering module. From left to right: full model; w/o register-based filter; w/o AH filter; w/o FP-GP filter. Query token for similarity calculation is marked with the red dot.}
   \vspace{-0.4 cm}
   \label{fig:abla_msk}
\end{figure}

\noindent \textbf{Effects of Spurious Token Filter.}~
The spurious token filter is a crucial component of UniRefiner, providing reliable supervision.~\cref{fig:abla_msk} summarizes quantitative and qualitative results from a progressive ablation of the filtering modules. Removing $\Gamma_{\text{reg}}$ (the register-based spurious detector) causes a performance drop, though the model remains functional. Removing the AH filter next allows tokens corresponding to the `bus' to dominate the road region. Finally, removing the FP--GP filter lets spurious tokens overwhelm most image tokens, producing severe segmentation failures—worse than the vanilla baseline. These findings confirm the need for a comprehensive spurious token filtering pipeline.

\begin{table}[t]
    \centering
    \caption{\textbf{Ablation on the designs of Register.} We ablate the key components of our register design and report the performance on ADE20K with linear probing.}
        \label{tab:ablation_register}
\vspace{-2 mm}
\begin{adjustbox}{width=0.42\textwidth}
\setlength{\tabcolsep}{12 mm}
\begin{tabular}{ccc}
\hline
Model & mIoU & mACC \\ \hline
\rowcolor[HTML]{EFEFEF}  Full Model & 49.8 & 62.9 \\ \hline
Learnable register & 47.9 & 60.1 \\
w/o/ $\gL_{\text{uni}}$ & 45.8 & 59.0 \\
w/o/ $\gL_{\text{spu}}$ & 45.0 & 58.5 \\ \hline
\end{tabular}
\end{adjustbox}
\vspace{-3 mm}
\end{table}

\noindent \textbf{Designs of Contrastive Register.}~
Since~\cite{chen2025vision} also employs post-hoc register designs, we conduct an ablation study that systematically removes our key innovations to revert to their baseline configuration: 1) replacing Gaussian-noise registers with learnable tokens; 2) removing uniformity loss $\gL_{\text{uni}}$ ; 3) removing spurious alignment loss $\gL_{\text{spu}}$. As shown in Table~\ref{tab:ablation_register}, each removal step progressively degrades performance, confirming the necessity of both register design and our contrastive learning objectives.

\begin{figure}[t]
  \centering
  \includegraphics[width=1.0\linewidth]{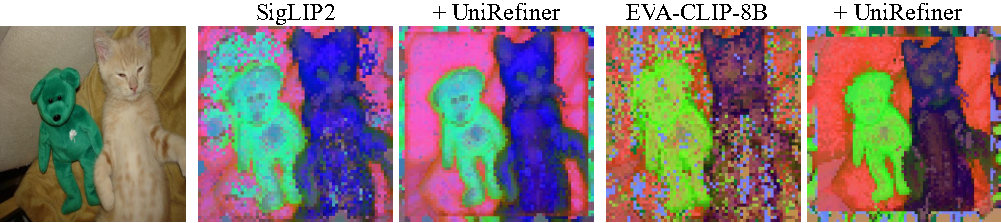}
  \vspace{-7 mm}
\caption{\textbf{PCA visualization comparison.} We compare the final-layer tokens of vanilla and refined SigLIP2-So400m and EVA-CLIP-8B. After refinement, register tokens on the image boundary absorb spurious tokens and redistribute them to the periphery, separating them from regular image tokens.}
   \label{fig:ana_reg}
   \vspace{-4.5 mm}
\end{figure}

\noindent \textbf{What is Learned in Registers?}~UniRefiner uses register tokens on the image boundary to absorb spurious tokens, thereby separating them from regular image tokens. To validate this behavior, we visualize final-layer token features of vanilla and refined models with PCA in~\cref{fig:ana_reg}. In the vanilla model, spurious tokens are widely mixed with image tokens. After refinement, these spurious tokens are redirected to the register region, while the image region is occupied by regular tokens. This visualization confirms that the learned registers effectively separate spurious tokens from the image region.



%% file: sec/7_conclusion.tex
\section{Conclusion}
\label{sec:conclusion}
We presented UniRefiner, a universal post-hoc refinement framework that effectively eliminates spurious tokens in pre-trained ViTs through contrastive register learning. By explicitly identifying and redistributing three types of spurious tokens, our method enhances spatial representation quality without structural modifications. Experiments demonstrate that UniRefiner unlocks the dense prediction potential of billion-parameter VLMs, enabling them to surpass specialized vision models across dense-level tasks, paving the way for their broader application.

\section*{Acknowledgements}
This work was supported by the National Natural Science Foundation of China under Grant No. 62376209.